\documentclass[paper]{ieice}

\usepackage{graphicx}
\usepackage{array}
\usepackage[fleqn]{amsmath}
\usepackage{newtxtext}
\usepackage[varg]{newtxmath}
\usepackage{cite}
\usepackage{multirow}
\usepackage{url}
\usepackage{amsmath,amssymb}
\usepackage{color}

\usepackage{soul}
\usepackage{pifont}
\usepackage[bottom]{footmisc}
\usepackage{hyperref}

\usepackage{silence} 
\usepackage{dblfloatfix}
\usepackage{subcaption}
\field{}
\title{}
\title{Cross-domain Deep Feature Combination for Bird Species Classification with Audio-visual Data}
\authorlist{%
	\authorentry[naranchimeg@scv.cis.iwate-u.ac.jp]{Bold Naranchimeg}{n}{labela}
	\authorentry[zhang@u-fukui.ac.jp]{Chao Zhang}{n}{labelc}
	\authorentry[akashi@iwate-u.ac.jp (Corresponding author)]{Takuya Akashi}{n}{labela}
}
\affiliate[labela]{Graduate School of Engineering, Department of Design and Media Technology, Iwate University, Iwate, 020-8551 Japan}
\affiliate[labelc]{Graduate School of Engineering, Information Science, University of Fukui, Fukui, 910-8507 Japan}

\begin{document}
\maketitle
\begin{summary}
	In recent decade, many state-of-the-art algorithms on image classification as well as audio classification have achieved noticeable successes with the development of deep convolutional neural network (CNN). However, most of the works only exploit single type of training data. In this paper, we present a study on classifying bird species by exploiting the combination of both visual (images) and audio (sounds) data using CNN, which has been sparsely treated so far. Specifically, we propose CNN-based multimodal learning models in three types of fusion strategies (early, middle, late) to settle the issues of combining training data cross domains. The advantage of our proposed method lies on the fact that We can utilize CNN not only to extract features from image and audio data (spectrogram) but also to combine the features across modalities. In the experiment, we train and evaluate the network structure on a comprehensive CUB-200-2011 standard data set combing our originally collected audio data set with respect to the data species. We observe that a model which utilizes the combination of both data outperforms models trained with only an either type of data. We also show that transfer learning can significantly increase the classification performance.  
\end{summary}


\maketitle

\section{Introduction}
\label{sec:intro}
Identifying the species of a bird is a widely-studied problem to ornithologists, and an important task in ecosystem monitoring and biodiversity preservation. Recognition of bird species in images is a challenging task due to various appearances, backgrounds, and environmental changes. Despite of this, this fine-grained recognition task has received a significant amount of attention in the computer vision community \cite{zhang2014part, branson2014bird, cui2016fine, gavves2015local} because of its potential widespread applications. Compared to generic object recognition, fine-grained recognition benefits more from learning critical parts of the object that can help align objects of same class and discriminate between neighboring classes. Current state-of-the-art methods (e.g., \cite{zhang2014part,branson2014bird,guo2018fine, lebedev2018impostor,he2017fine}) adopt CNN-based architectures that learn representations directly from the raw data and can be used to extract set of discriminative features.
\par
On the other hand, sound also provide us important information about the world around us. Many animals make sounds either for communication or their living activities such as moving, flying, mating etc. Although sound is in some case complementary to visual information, such as when we listen to something out of view, vision and hearing are often informative about the same structures in the world \cite{owens2016ambient}. As a consequence, numerous efforts have been devoted to recognize bird species based on auditory data \cite{kahl2017large,cakir2017convolutional} in recent years. Adapting CNN architectures for the purpose of audio event detection has become a common practice and generating deep features based on visual representations of audio recordings has proven to be very effective \cite{takahashi2018aenet} such as in bird sounds \cite{piczak2016recognizing,cakir2017convolutional}. To improve the affinity between images and sounds, we utilize the CNN to extract features from spectrogram of audio recordings.
\par 
In the real world, human are able to handle information consist of different information from multiple plural modalities. In the filed of machine learning, multimodal recognition can improve performance compared with unimodal recognition by utilizing complementary sources of information \cite{srivastava2012multimodal,ngiam2011multimodal}. Multimodal learning have been used for tasks such as image and sentence matching \cite{tatulli2017feature}, RGB-D object recognition \cite{eitel2015multimodal}, action detection \cite{simonyan2014two} and specially speech recognition \cite{tatulli2017feature,noda2015audio,meutzner2017improving,torficoupled}, fusing different modalities. The results have shown that one modality can enhance the performance of the other by providing relevant information. Furthermore, authors of \cite{ngiam2011multimodal,huang2013audio} proposed fusion schemes for multimodal learning with considering the architectures of neural networks. 
\\
Our main contributions are summarized as follows:
\begin{itemize}
    \item We propose that the combination of image and sound provide richer training signal for bird species classification under CNN framework, which is the first attempt to the best of our knowledge.
    \item Three strategies are investigated for fusing audio and image modalities using CNN.
\item We collect at least 10 audio recordings for each bird over 178 species, corresponding to the image dataset CUB-200-2011 \cite{wah2011caltech}.
\end{itemize}
Specifically, we adopt CNN to process jointly the two modalities for bird species classification. Three strategies are investigated for fusing audio and image modalities using CNN: (1) an early fusion strategy in which the feature vectors related to each modality are concatenated together and input to the CNN. (2) A middle fusion strategy. Features learned by each single modality are combined at the mid-level of the CNN. (3) A late fusion strategy. Outputs of single modality are fused to determine a final classification. 
Our experimental results show that the architecture with late fusion strategy outperforms among the proposed architectures, which indicates that combining decisions of the classifiers from two modalities is superior. In addition, we apply a two-stage training procedure, which improves the classification accuracy. 
\par
The rest of the paper organized as follows. In section 2, we review related works on bird species identification with CNN and multimodal learning algorithms. Section 3 describes several architectures among which a multimodal CNN processing jointly the two modalities. In section 4, we describe the integrated dataset utilized for the evaluation experiments of our architectures. Section 5 describes the experimental setup and results. In section 6, we conclude this paper. 

\section{Related work}
\label{sec:re}
\par
Our work is related to the problem of recognition from multimodal data as well as convolutional neural networks for image classification. We will briefly highlight connections and differences between our work and existing works. 
\par
Deep neural networks (DNN) have successfully applied for single modality such us text \cite{collobert2011natural,zhang2015character,kim2014convolutional}, images \cite{le2013building,lecun1990handwritten,lecun1989backpropagation} and audio \cite{hinton2012deep,petetin2015deep} showing their ability to learn representations directly from raw data and can be used to extract a set of discriminative features. CNN is one powerful deep architecture of DNN commonly utilized for image classification \cite{krizhevsky2012imagenet, szegedy2015going, he2016deep}. The use of CNN for distinguishing between fine grained categorization such as bird species categorization has been proposed in many studies \cite{zhang2014part,branson2014bird,guo2018fine,lebedev2018impostor,he2017fine}, which employ part/pose based approaches to achieve good performance. Besides, CNNs have also been applied for speech processing \cite{sainath2013deep,abdel2014convolutional}. In \cite{takahashi2018aenet,espi2015exploiting}, authors utilized CNNs to extract features from spectral representations of audio recordings. It has proved to be efficient when extracting features based on spectrograms of audio recordings such as bird sound \cite{kahl2017large,piczak2016recognizing}, thus we used spectrogram representations for audio data. 
In this paper, instead of employing part/pose based approaches we will focus on integrating raw image an audio using conventional CNN, in order to know which fusion strategy performs best. 	 
\par 
Multimodal learning algorithms have been used for tasks such as image sentence matching \cite{ma2015multimodal}, action recognition \cite{simonyan2014two}, RGB-D object recognition \cite{eitel2015multimodal}, and speech recognition (audio-visual speech recognition \cite{noda2015audio} and visual-only speech recognition \cite{tatulli2017feature}). Among the many approaches for multimodal learning, multimodal integration is commonly realized by three different categories of approaches. First, in early fusion approach, feature vectors from multiple modalities are concatenated and transformed to acquire a multimodal feature vector. For example, Ngiam et al. \cite{ngiam2011multimodal} utilized DNN to extract fused representations directly from multimodal signal inputs. Likewise, in middle fusion approach, Huang et al. \cite{huang2013audio} employed deep belief network (DBN) to combine mid-level features learned by single modality. Lastly, in late fusion approach, outputs of unimodal classifiers are merged to determine a final classification. For example, in RGB-D object recognition, Eitel et al. \cite{eitel2015multimodal} proposed two separate CNN streams processing RGB and depth data independently are combined with late fusion approach. Therefore,  Simonyan et al. \cite{simonyan2014two} proposed two-stream (one stream processing spatial features from RGB image inputs, while the other stream processing temporal features from optical flow inputs) network architecture designed to recognize action for videos. They combined two streams by concatenating features and by averaging prediction scores from two CNNs, respectively. In contrast to these works, we propose simple yet effective concatenation, summation and multiplication based fusion methods with respect to three strategies.

\section{Methodology}
\label{sec:method}
Our three multimodal architectures extend conventional CNN for large-scale image classification \cite{krizhevsky2012imagenet}. Our implementation is based on CaffeNet , and can be treated as \cite{jia2014caffe} a variation of the structure proposed by Krizhevsky et al. \cite{krizhevsky2012imagenet}. 
\subsection{Feature extraction}
CNN uses hierarchical features in its processing pipeline. The features from initial layers are primitive while late layers are high-level abstract features made from combinations of lower-level features. The CaffeNet consists of five convolutional layers (with max pooling layers following the first, second and fifth convolution layer) followed by three fully connected (FC) layers and a softmax classifier. Rectified linear unit is applied to every convolutional layer and fully connected layer and  local normalization is applied in first and second convolutional layer. The process through this 8-layer CNN network can be treated as a process from low to mid to high-level features. We hypothesize, that combining the features of different layers in this pipeline can lead to achieve better performance. 
\subsection{Feature combination}
We propose our method in three strategies to fuse features: early fusion, middle fusion and late fusion. Early fusion, also known as feature level fusion, is a feature combination scheme that features from multiple modalities concatenated to form a merged feature vector. Middle fusion, also called mid-level combination, combines the high-level features learned by single network. Late fusion, also called decision-level fusion, combines the decisions of the unimodal classifier and determine the final classification. In this paper, we use concatenation to combine low and high-level features, and summation or multiplication to combine decisions of the classifiers. 
The feature combination layers can be trained with standard back-propagation and stochastic gradient descent.   

\begin{figure}[tb]
\includegraphics[width=\linewidth]{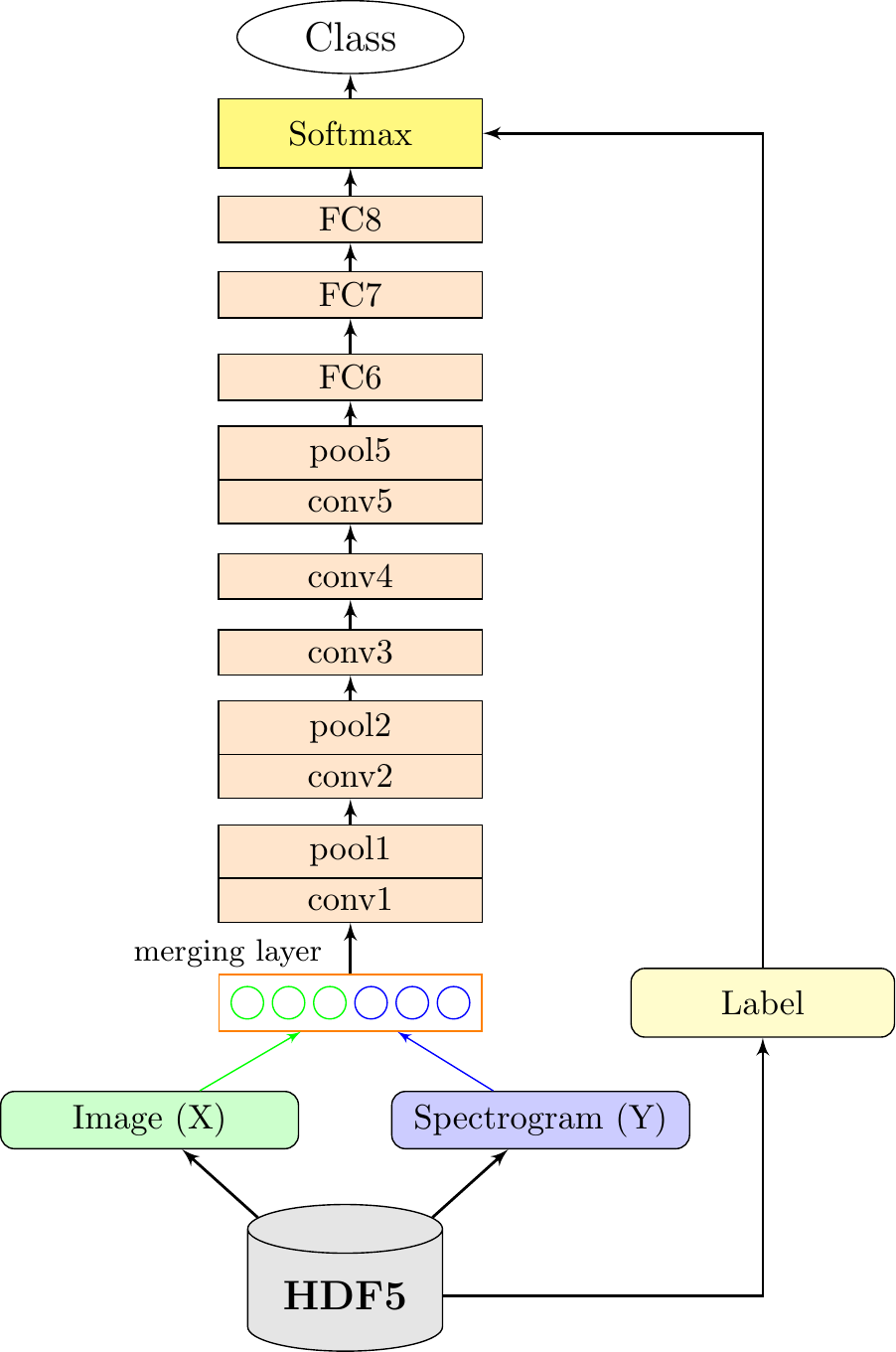}
\caption{The architecture of early fusion model (Net1). $ 227 \times 227$ pixel RGB images of two modalities are concatenated at merging layer, which produces $ 227 \times 454 \times 3$ output volume, and the convolution layers will extract joint features from this merged volume. We use HDF5 format to manage datasets of two modalities, because of it's flexible data storage and unlimited data types.}
\label{net1}
\end{figure}

\subsection{Architectures of proposed models}
\label{sec:apm}

The proposed multimodal learning models which combines audio and image using CNN by different fusion approaches are described in this section. We exploit the same architecture for both audio and image modalities to focus on evaluating the effectiveness of the feature combination approaches.
\subsubsection{Early fusion model}
One direct approach for combining audio and image is to train a CNN over the concatenated audio and image data as shown in Fig. \ref{net1}. In this strategy, the input vectors related to each modality are concatenated together and then processed together throughout the rest of the CNN pipeline. This model is most computational efficient comparing to the middle and late fusion models, because the number of learnable weight parameters is almost half times less than late fusion model.

\subsubsection{Middle fusion model}
\begin{figure}[tb!]
	\includegraphics[width=\linewidth]{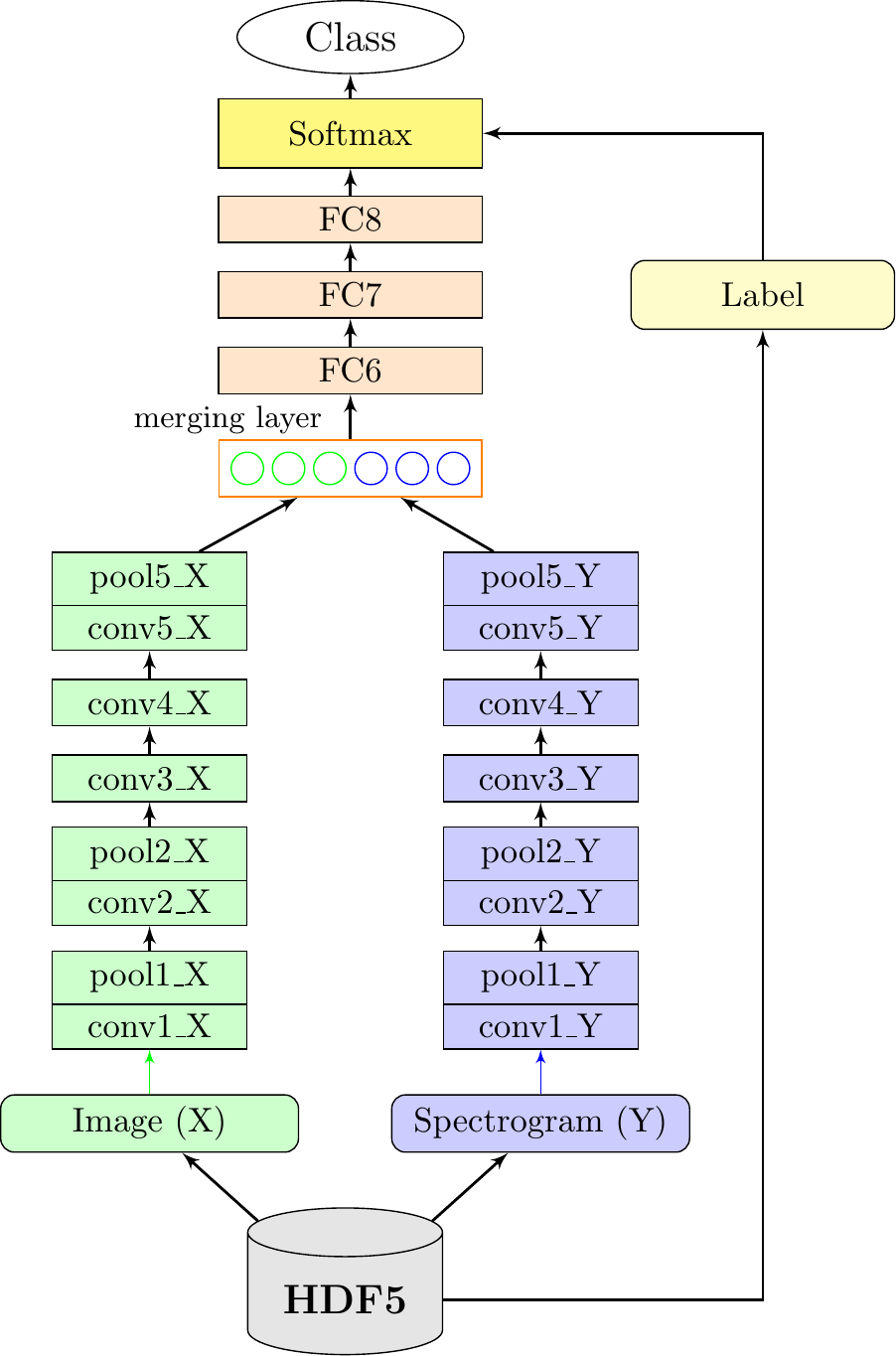}
	\caption{The architecture of middle fusion model (Net2). The activations of the pool5 layers of the two modalities are concatenated at merging layer, and feeding it into the three fully connected layers with softmax at the end. }
	\label{net2}
\end{figure}

In the middle fusion strategy, unimodal features is extracted independently from audio and images, then combined into a multimodal representation by concatenating the activations of the last pooling layers of two modalities. The multimodal representation then learned in the following fully connected layers. The middle fusion model is shown in Fig.\ref{net2}.

\subsubsection{Decision or late fusion model}
\begin{figure}[t]
	\includegraphics[width=\linewidth]{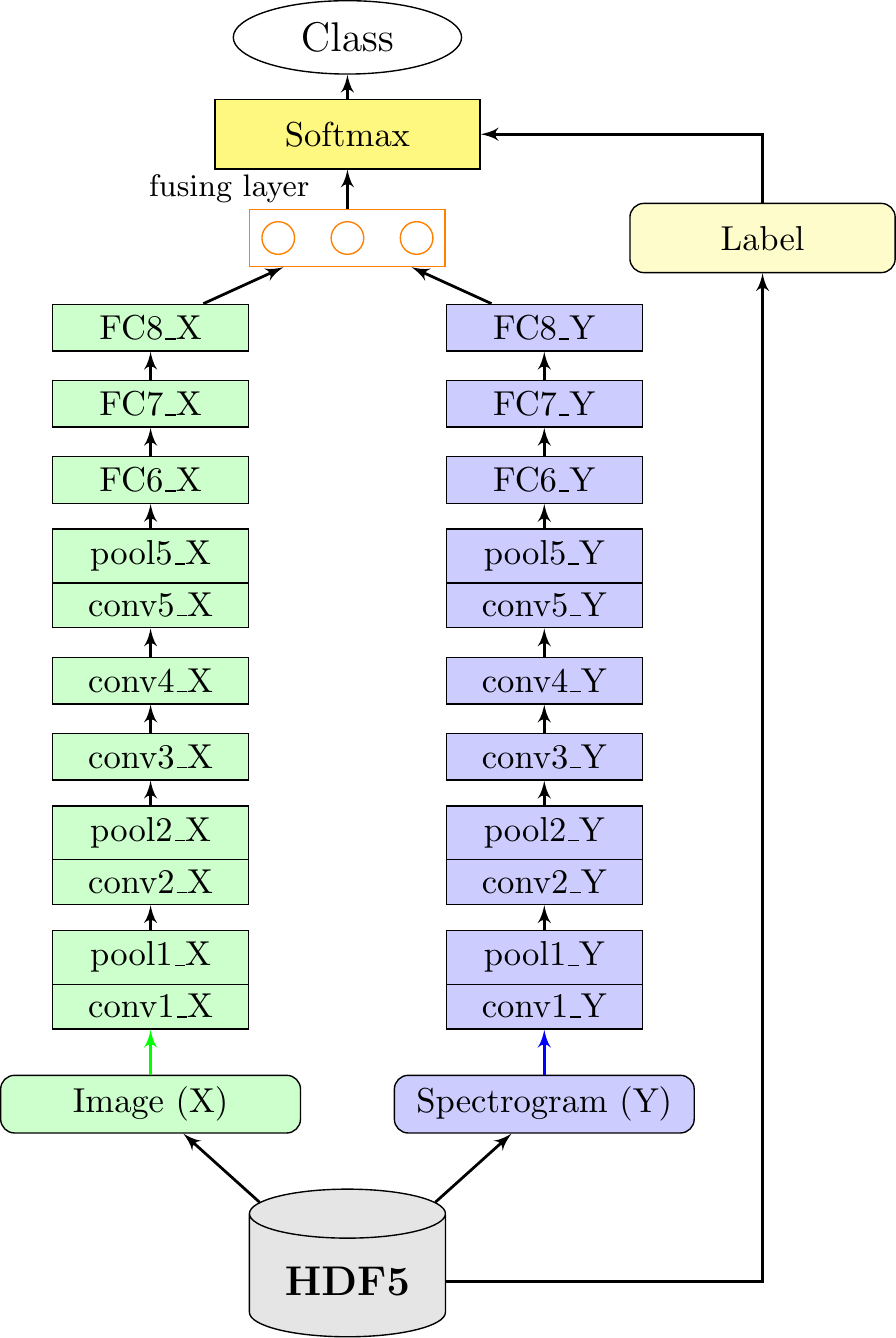}
	\caption{The architecture of late fusion model (Net3). The last fully connected layers of each model holds the unimodal scores for each class, which fused at fusing layer by summing and multiplying the unimodal scores. }
	\label{net3}
\end{figure}

In contrast to the middle fusion model, extracted unimodal features are separately learned to compute unimodal scores, then these scores are integrated to determine a final scores. The late fusion model consists of two-streams processing audio and image data (green and blue) independently, which are fused after last fully connected layers as shown in Fig.\ref{net3}. Among the various ways of combining CNNs with different modalities, one straightforward way is to add additional fully connected layers to combine the outputs of the each streams as presented in \cite{eitel2015multimodal}. Instead of using FC layer to combine the two streams, we applied element-wise summation and element-wise multiplication to fuse the outputs of the each streams. 

\section{Database}
\label{database}
We evaluate our models on the popular fine-grained CUB-200-2011 \cite{wah2011caltech} bird dataset and our originally collected sound dataset from sharing bird sound database Xeno-Canto\footnote{\url{http://www.xeno-canto.org}}.

\begin{figure}[t]
\centering
\begin{subfigure}[b]{0.45\textwidth}
   \includegraphics[width=\linewidth]{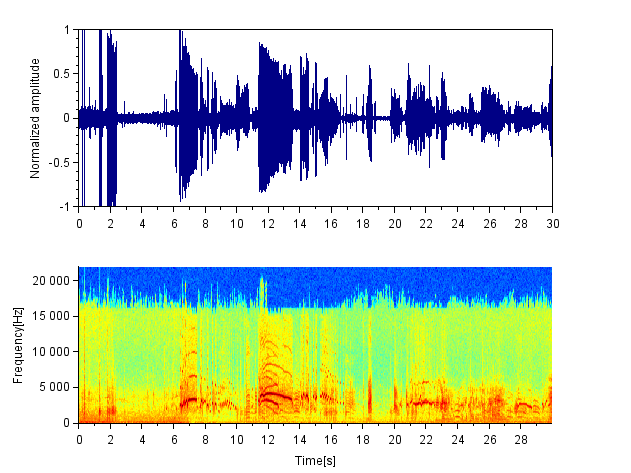}
   \caption{}
\end{subfigure}
\begin{subfigure}[b]{0.4\textwidth}
   \centering {\includegraphics[width=0.8\linewidth]{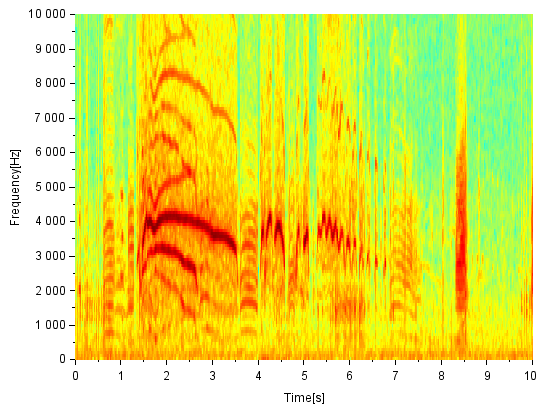}}
   \caption{}
\end{subfigure}
\caption[Spectrogram of the black-footed albatross]{(a) Radio (top) and spectrogram (bottom) of the black-footed albatross. (b) The spectrogram of 10 seconds duration, which will be fed into the bird classification model.}
\label{spectrogram}
\end{figure}

\begin{figure}[t] 
	\centering
	\begin{subfigure}{.25\linewidth}
		\centering
		\includegraphics[width=0.95\textwidth]{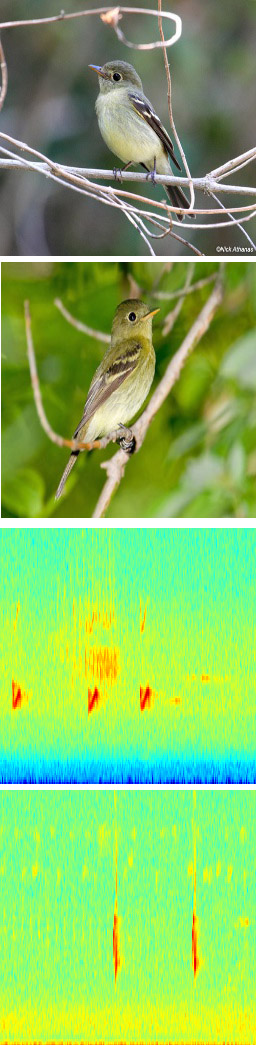}
		\caption{}
	\end{subfigure}%
	\begin{subfigure}{.25\linewidth}
		\centering
		\includegraphics[width=0.95\textwidth]{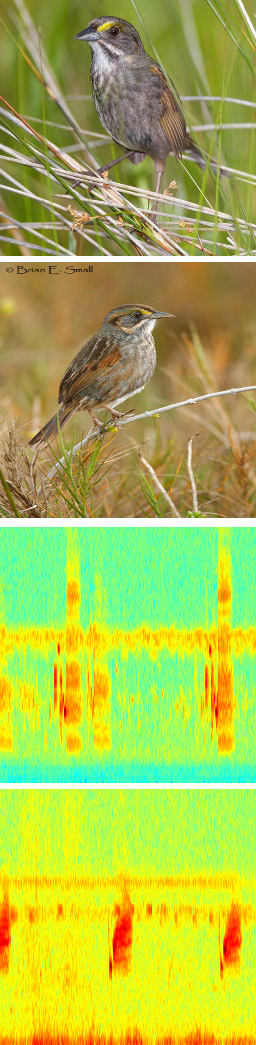}
		\caption{}
	\end{subfigure}%
	\begin{subfigure}{.25\linewidth}
		\centering
		\includegraphics[width=0.95\textwidth]{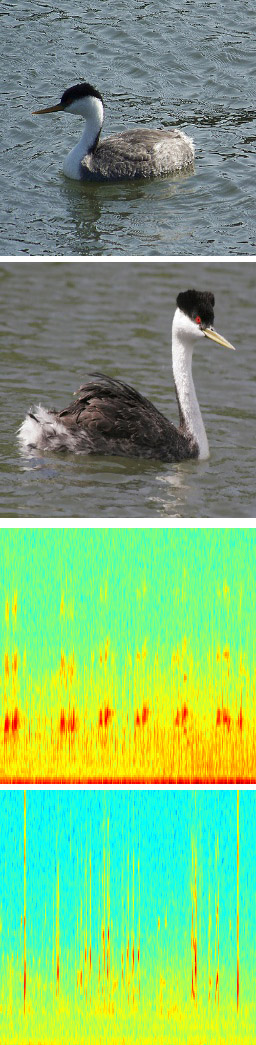}
		\caption{}
	\end{subfigure}%
	\begin{subfigure}{.25\linewidth}
		\centering
		\includegraphics[width=0.95\textwidth]{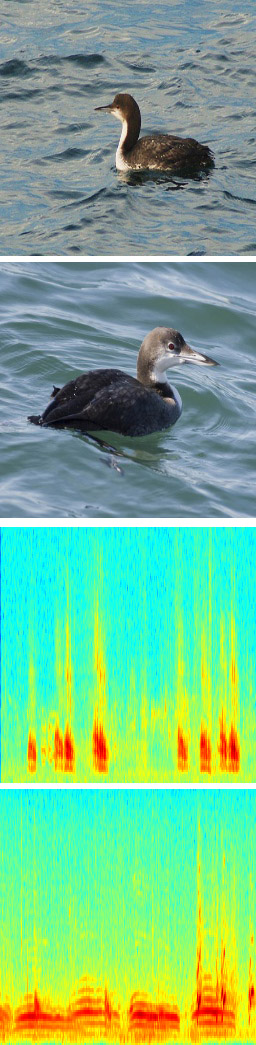}
		\caption{}
	\end{subfigure}%
	\caption{An example of CUB-200-2011 and audio dataset: (a) the yellow bellied flycatcher, (b) the seaside sparrow (c), the western grebe, and (d) the pacific loon.}
	\label{dataset}
\end{figure}

CUB-200-2011 \cite{wah2011caltech} dataset contains 11788 images of 200 species of birds, with each image downscaled to 227 $\times$ 227 pixel. Spectro-temporal features (spectrogram) are extracted from audio recordings that we collected from the Xena-Canto to be used as the audio representation. Based on the 200 species of the CUB-200-2011, we try to harvest at least 10 different audio recordings for each species. As a result, audio recordings over 178 species were collected completely ($\mathrm{\#recordings}\geq10$), audio recordings from 19 species were collected deficiently ($0<\mathrm{\#recordings}<10$), and 3 species could not be collected. The spectrograms of the audio are obtained by using short-time fourier transform (STFT) over 10 seconds audio frames, windowed with Hanning window (size 512, 50\% overlap). The reason is, the sounds of birds are usually contained in a small portion of the frequency range (mostly around 2-8 kHz) as stated in \cite{cakir2017convolutional}, so we only extract features from the range of (0, 10) kHz. In order to focus only sounds produced in the vocal organ of birds (i.e. calls and songs), first we obtained the maximum amplitude of the audio and removed a frame which contains only amplitude less than $1/4$ of the maximum amplitude. Finally, the spectrograms are saved as 227$\times$227 pixel color images, and the dataset contains 4807 images of 194 species of birds. 
Several examples including both images and spectrograms over different bird species are shown in Fig.\ref{dataset}. We follow the standard training/test split of CUB-200-2011 dataset suggested in \cite{wah2011caltech}. The sound dataset is split into two halves for training and test set respectively. 
\par
The multimodal CNNs are trained in a supervised manner, thus we create integrated dataset by matching two data sets and corresponding labels using HDF5 \footnote{\url{https://www.hdfgroup.org/HDF5/}} file format. Because the HDF5 file can contain any collection of data entities (images, videos, audio recording, text, etc.,) in a single file, and used to organize  heterogeneous collections consisting of very large and complex datasets. 

\section{Experiments and Results}
\label{experiment}
The experiments in this section are conducted to evaluate the effectiveness of our proposed architectures (Net1, Net2, Net3). In order to evaluate an advantage of our late fusion approach, we conduct a comparative experiment between Net3 model and two different existing fusion approaches \cite{eitel2015multimodal,simonyan2014two}. Besides, we have fine-tuned a pretrained model for all the models (including comparative models) in order to improve the performance. To improve the repeatability and only focus on the evaluation of fusion step, we use the well-known CaffeNet model \cite{jia2014caffe} to extract features, train, and fine-tune the CNN with default structure and parameter setting fixed except the base learning rate and batch size. We set the learning rate to 0.001 for single modality and 0.0001 for multimodal learning. The batch size is set to 32 for single modality and 1 for multimodal learning due to limited resources. 

\subsection{Quantitative result: Single-modality v.s. Multi-modality}
The core idea of this paper is to address the integration of the image and audio. Therefore, it is necessary to compare the performance between single-modality (image or sound) and multi-modality (both image and sound: Net1, Net2, Net3). To focus on the evaluation of fusion models, in this experiment, we do not introduce any transfer learning techniques (e.g., fine-tune the pretrained model to help differentiate between gains from the proposed architectures). Table \ref{table1} summarizes the results of single modality and proposed models. We can observe that combining two modalities using CNN improves the performance of those only using one modality, extracting features separately from image and audio and fusing them at the late stage performs better with significant gain. Interestingly, the performance of low-level and mid-level fusion models slightly better than the performance of single-modality. One possible reason is because CNN learns features for the predominant modality. In contrast, learning features separately for different modalities results in more independent features, which leads to achieve better performance. Let us mention that the result obtained by multi-modality is different from simply combining the results of two CNNs trained separately. Indeed, the two modalities' parameters are jointly estimated and thus can be mutually influenced. We provide a detailed discussion in following section. 
\begin{table}[!t]
	\caption{Comparative results between individual modality and multimodal CNNs.}
	\label{table1}
	\renewcommand{\arraystretch}{1.2}
	\centering
	\begin {tabular}{cl|c}
	\hline \hline%
	
	\multicolumn {2}{c}{Method} & \parbox{2cm}{\centering Accuracy (\%)}
	\\\hline
	\multirow{2}{*} {Single modality} 		& Image	& 16.2		\\\cline{2 - 3}
	& Audio & 46.4	\\\hline
	\multirow{3}{*} {Multi modality} 	& Net1 	& 50.0		\\\cline {2 -3}
	& Net2 	& 49.9		\\\cline {2 -3}
	& \bf Net3 (summation)	& \bf 53.8	\\\hline
	\end {tabular}
\end{table}

\begin{figure}[b]
	\centering
	\includegraphics[width=\linewidth]{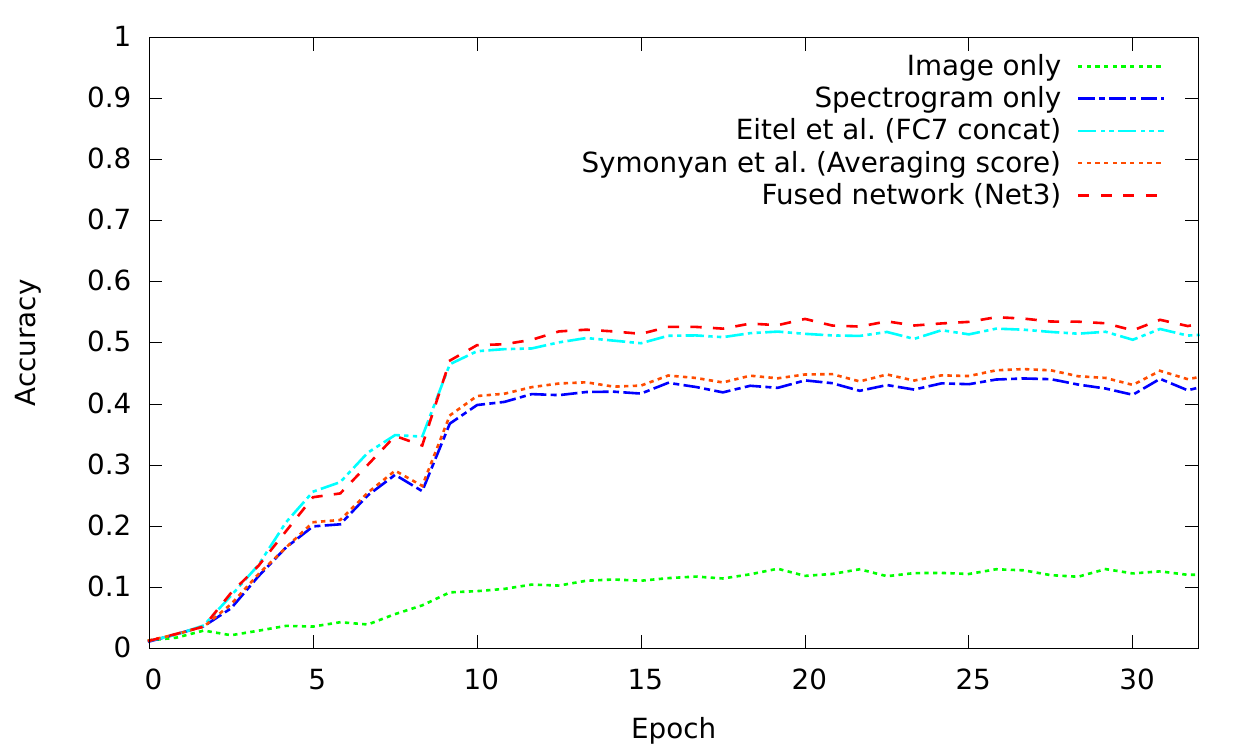}
	\caption{Test accuracy vs. Epoch.}
	\label{net3_all}
\end{figure}

Figure \ref{net3_all} plots learning process of single modality (image only and audio only) and Net3 model over learning epochs. It can be observed that Net3 significantly improves the results.

\subsection{Qualitative result: Single-modality v.s. Multi-modality} \label{ssec:qualitative}
\begin{figure}[t]
\centering
\begin{tabular}{m{2.5em}ccc}
\multirow{2}{*}{\rotatebox[origin=c]{90}{Input}} & \includegraphics[width=0.23\linewidth]{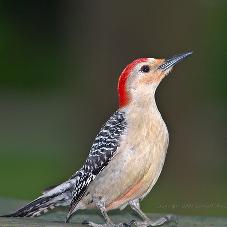}& \includegraphics[width=0.23\linewidth]{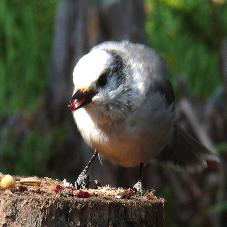}& \includegraphics[width=0.23\linewidth]{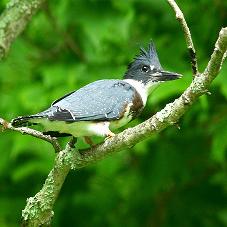}
\\
& \includegraphics[width=0.23\linewidth]{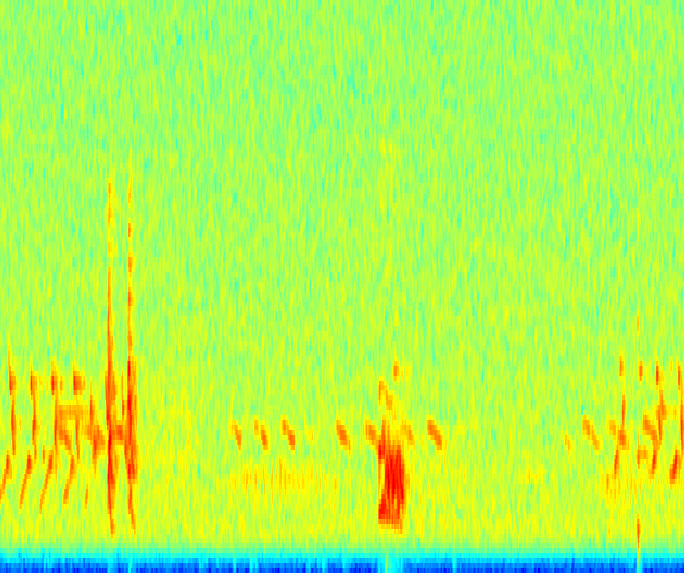} & \includegraphics[width=0.23\linewidth]{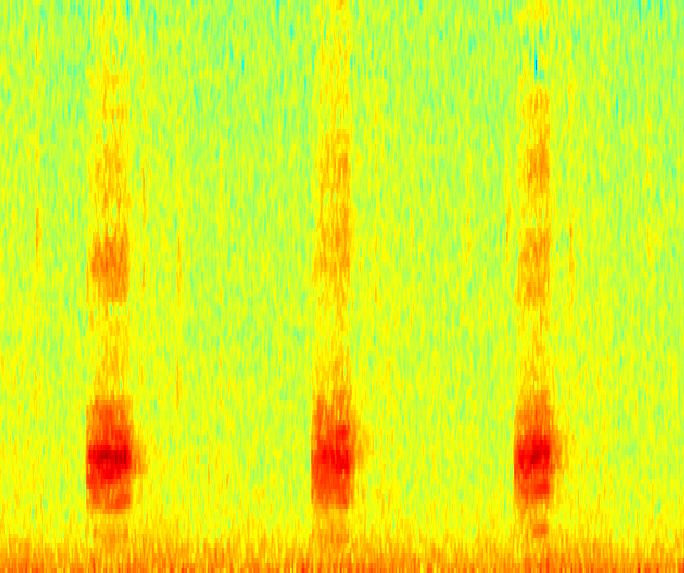} & \includegraphics[width=0.23\linewidth]{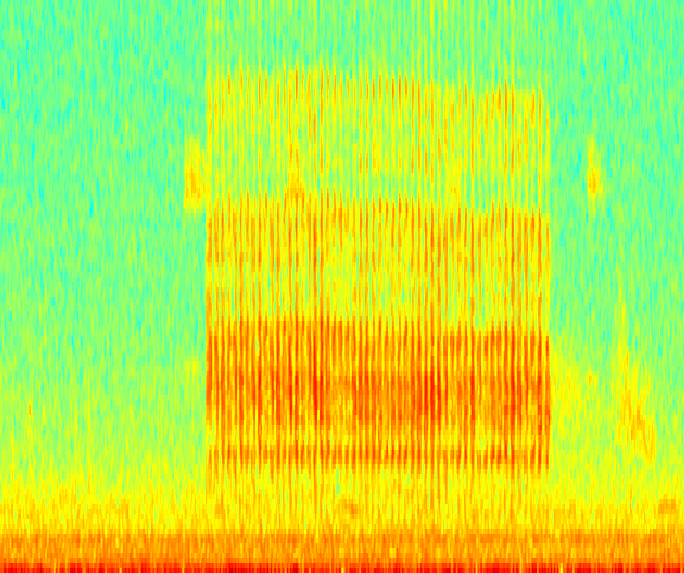} 
\\
Image & \ding{55} &  \ding{55}  &  \ding{55}\\ 
Sound & \ding{55} &  \ding{51} &  \ding{55}\\ 
Net1 & \ding{51} & \ding{51} & \ding{55}\\ 
Net2 & \ding{51} & \ding{51} & \ding{55}\\ 
Net3 & \ding{51} & \ding{51} & \ding{51}\\ 
\end{tabular}
\caption{Effects of combining image and spectrogram. Top two rows show sample image and spectrogram of different bird species where are fed into single-modality models and multi-modality models. The bottom rows show the resulting classification, where multimodal networks provide a correct classification while the single-modality classification are incorrect.}
\label{singlevsmulti}
\end{figure}

We perform a qualitative study to analyze effects of multimodal learning models by comparing single-modality and multi-modality networks. First, we select some classes where the multi-modality models provides the correct answer while the single-modality model produces the wrong classification. Then, we study why the multi-modality models provides right answers while the single-modality model failed to produce the correct classification. Figure \ref{singlevsmulti} shows some examples of single-modality vs. multi-modality classification. In the first column, the single-modality models predict the input image and spectrogram as the 'barn shallow' and the 'ring-billed gull' respectively rather than the 'red-bellied woodpecker'. However, the multi-modality models are able to predict the right answer, because those models provide joint features of different modalities. We observe that when single-modality model provide right answer for spectrogram, the probability of providing the right answer of multimodality models is higher than when single-modality classificaion is correct for image. In the second column of Fig. \ref{singlevsmulti}, single-modality model is misclassified the 'clark nutcracker' image as the 'great grey shrike', where other models provide correct answers. Lastly, the Net3 model is able to provide right answers while the other models provide misclassification on the 'belted kingfisher' (last column).
\par
\begin{figure}[!b]
\begin{tabular}{m{1em}m{3cm}m{3cm}}
\rotatebox[origin=c]{90}{Single-modality} &\includegraphics[width=\linewidth]{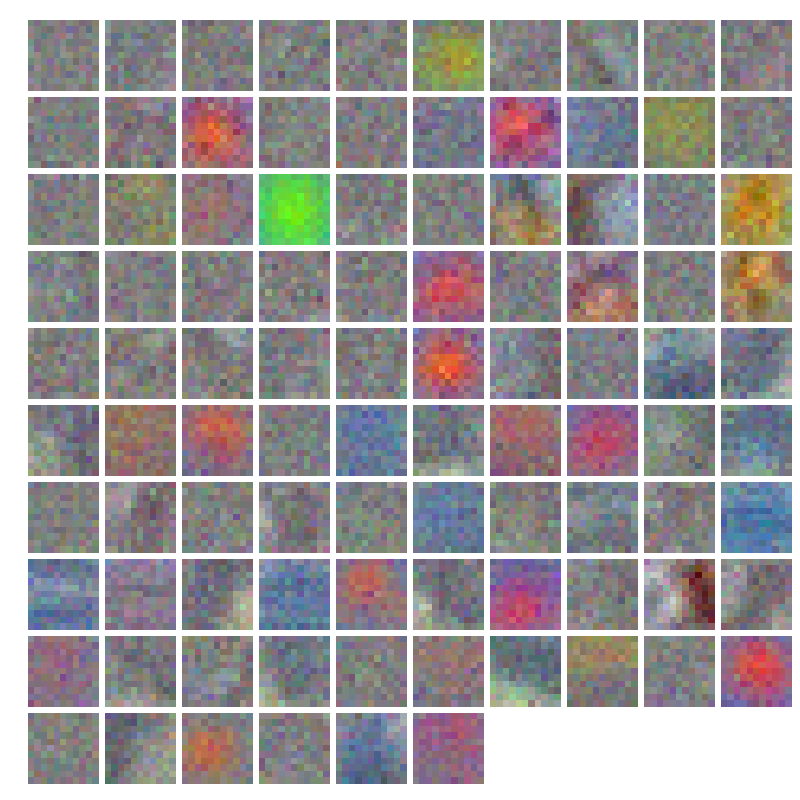}&
\includegraphics[width=\linewidth]{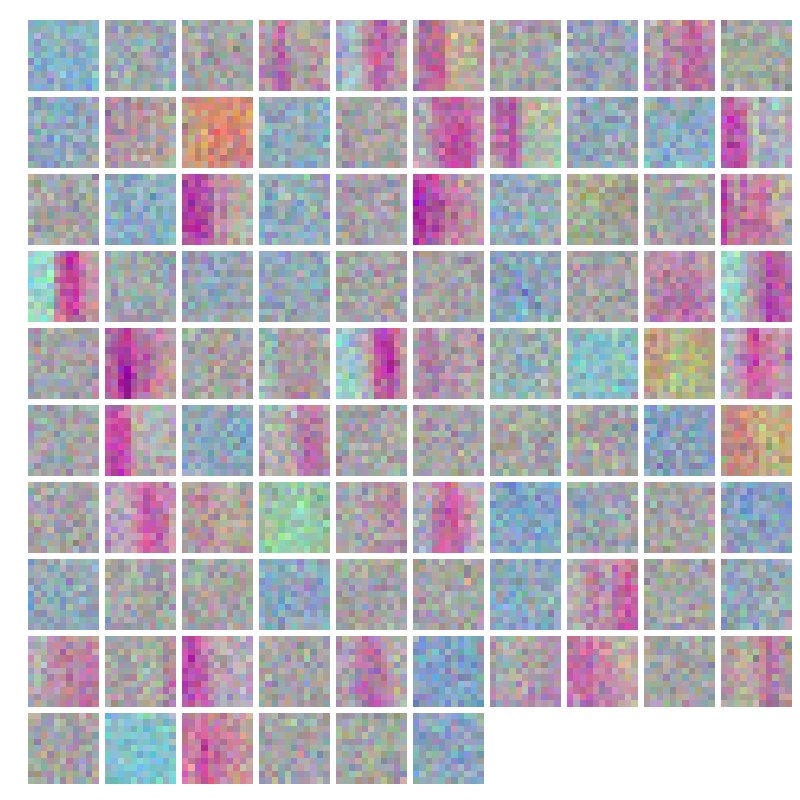} \\
\rotatebox[origin=c]{90}{\centering{Net1}} & \multicolumn{2}{c}{\includegraphics[width=.38\linewidth]{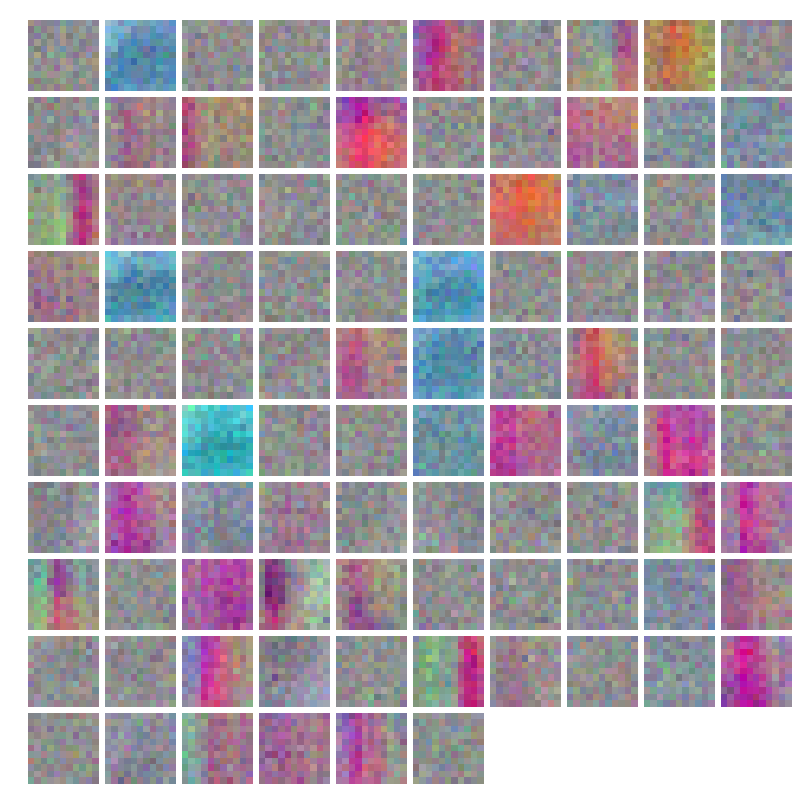}} \\
\rotatebox[origin=c]{90}{Net2} &\includegraphics[width=\linewidth]{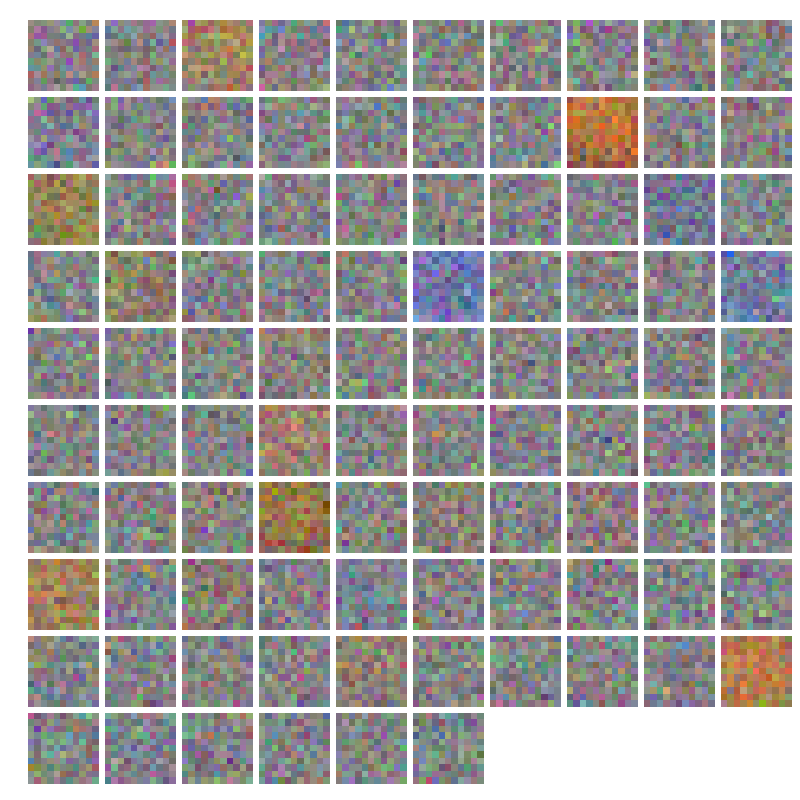}&
\includegraphics[width=\linewidth]{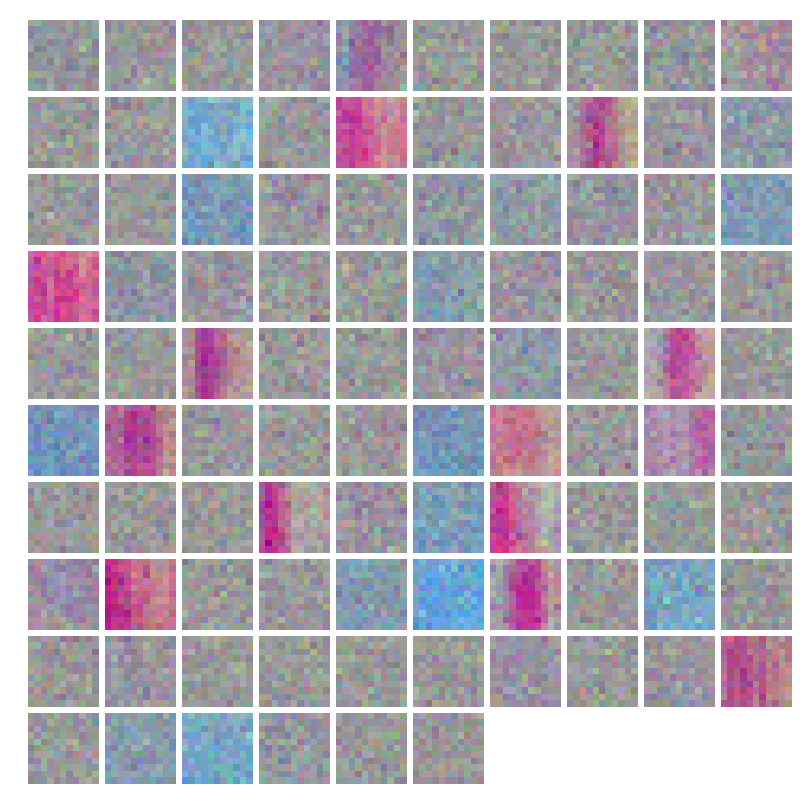}\\
\rotatebox[origin=c]{90}{Net3} &\includegraphics[width=\linewidth]{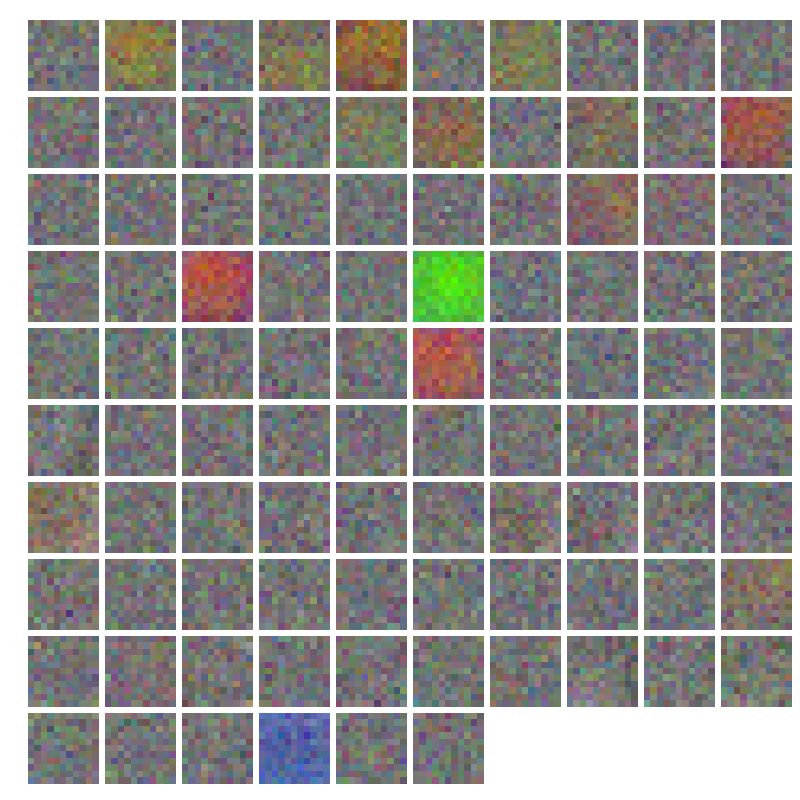} &
\includegraphics[width=\linewidth]{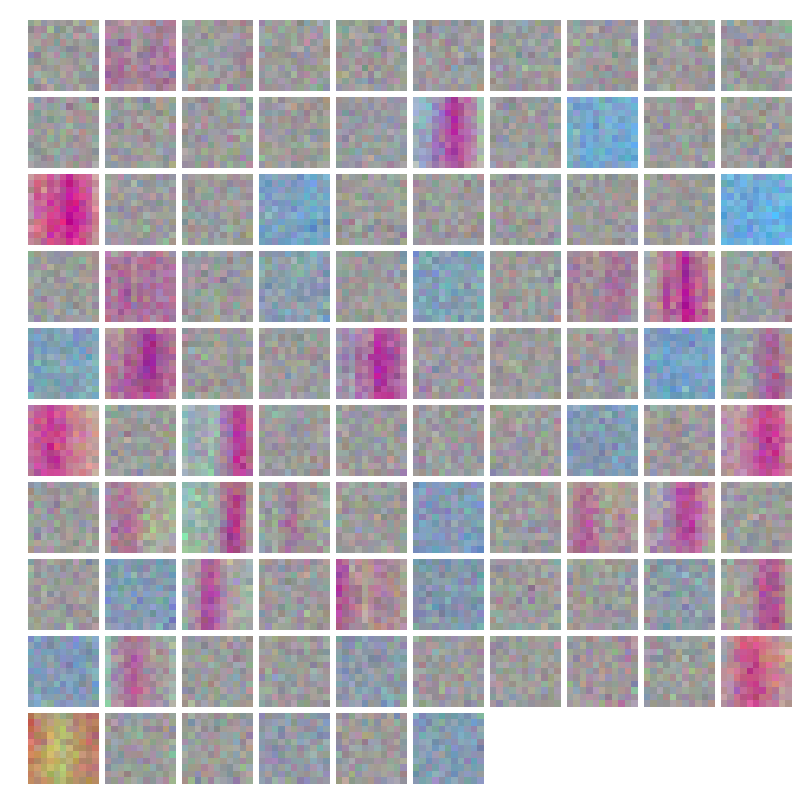}
\end{tabular}
\caption{Visualization of 96 filters of the first convolutional layer. Left side shows the filters related to the image network, while right side shows the filters related to the spectrogram network. It can be seen that the filters (left or right side) of different models have similar pattern. However, the filters of Net1 seems to have mixed filters of both networks.}
\label{filter}
\end{figure}

To better understand the difference between the models, we analyze the feature learned from the each networks by visualizing the filters of the first convolutional layers shown in Fig. \ref{filter}. We see that each network's filters of different models have similar pattern. Secondly, we see that the single-modality's filters have more meaningful patterns than the multi-modality filters. As we mentioned before its seems that the learning features separately for different modalities results more independent features. Finally, it can be seen that the filters of early fusion model has combined patterns from both networks, but most filters are similar to the spectrogram network. It reveals that the multimodal model (early fusion and mid-level fusion) learns features for the predominant modality. 

\subsection{Quantitative result: Net3 v.s. existing late fusion approaches}
\begin{figure}[b]
    \centering
    \includegraphics[width=.9\linewidth]{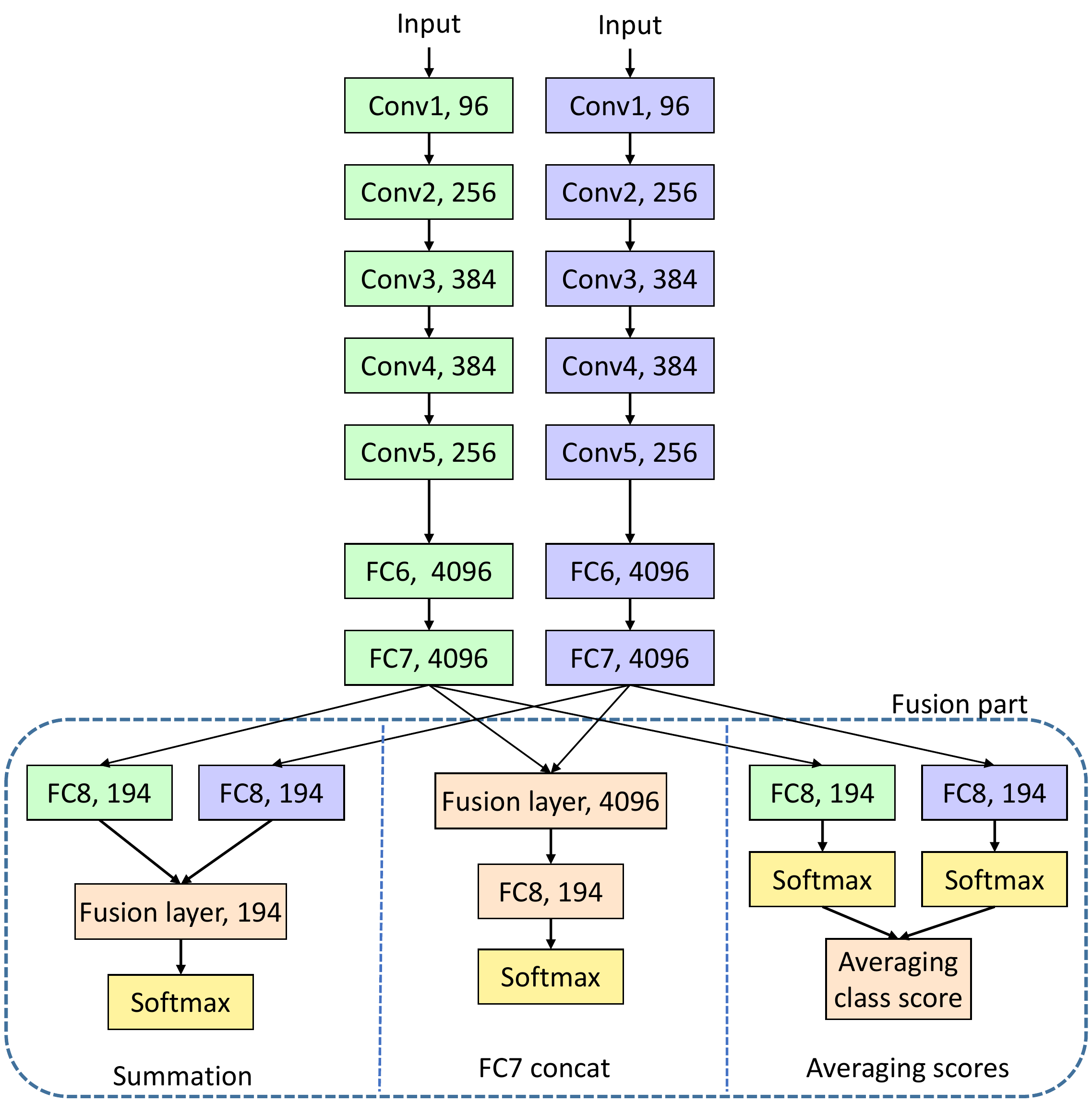}
    \caption{Differences between the late fusion approaches.}
    \label{fig:fusion_difference}
\end{figure}

To evaluate the effectiveness of our late fusing approach, we conduct comparative experiments on Net3 with existing late fusing methods presented in \cite{simonyan2014two,eitel2015multimodal}. The differences between the late fusion approaches is shown in Fig. \ref{fig:fusion_difference}.

\begin{itemize}
\item \textbf{FC7 concat\cite{eitel2015multimodal}}: FC7 layers (green and blue) of each networks are concatenated and merge into the fusion layer, which performs tensor multiplication of two vectors. The resulting fusion vector is then passed through one additional fully-connected layer for classification. This means this fusion methods is a linear combination of pair-wise interactions between two features. However, this method is not suitable when the features are in different sizes. 
\item \textbf{Averaging scores\cite{simonyan2014two}}: Each networks focus on learning features from images and spectrograms, respectively, and the final classification is computed as an average of the softmax scores of the two networks. In this fusion method, they do not consider pair-wise interactions between the features. However, this method is suitable when the model consists of different structured network streams. 

\end{itemize}
In terms of pair-wise interactions, our method is similar to FC7 concat method. However, we fused final output of each networks. Figure \ref{net3_all} plots learning curve of Net3 model with different late fusing methods for each epoch, indicating that averaging the Softmax scores gives lowest performance and the our fusing approach performs best.

\subsection{Fine-tuning the pretrained model}
Combining multi-modalities at the late stage of CNN has proven to be more effective, thus we show additional results with fine-tuning CaffeNet pretrained CNN under Net\ref{net3} model in this section. One natural idea for fine-tuning is to train the model by initializing both image and audio CNNs with the weights and biases of the first seven layers derived from CaffeNet pretrained network, discarding the last fully connected layer. Instead of last fully connected layer of the pretrained model, we randomly place the initialized new fully connected layer for 200-class bird classification (in our experiment, 194 classes due to the lack of audio dataset). 
\par
\begin{figure}[p]
\centering
\begin{tabular}{m{1em}m{3cm}m{3cm}}
\rotatebox[origin=c]{90}{Input} &\includegraphics[width=\linewidth]{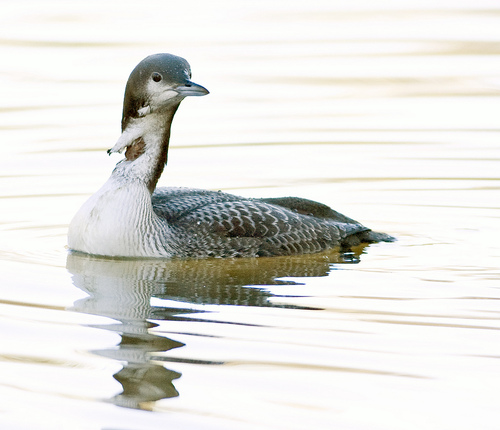} &
\includegraphics[width=\linewidth]{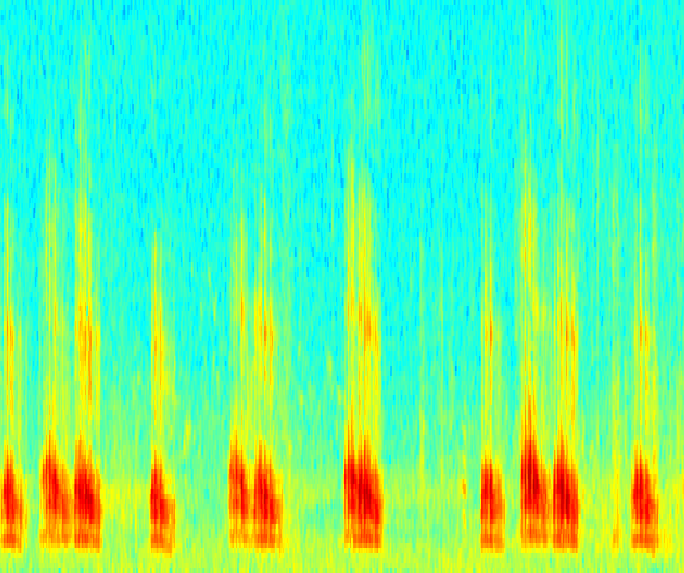} \\
\rotatebox[origin=c]{90}{Pool1} &\includegraphics[width=\linewidth]{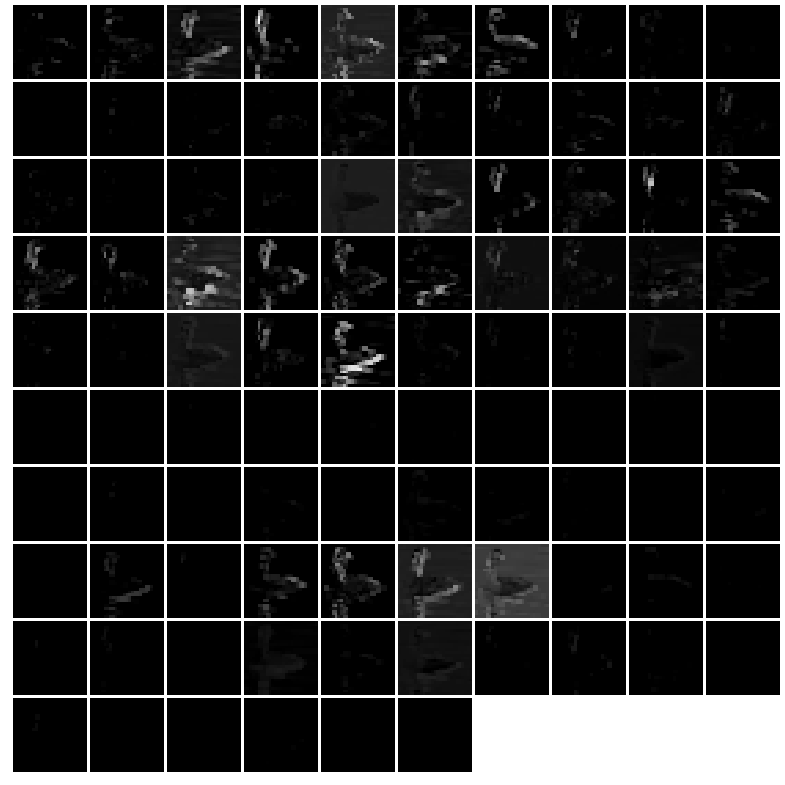} &
\includegraphics[width=\linewidth]{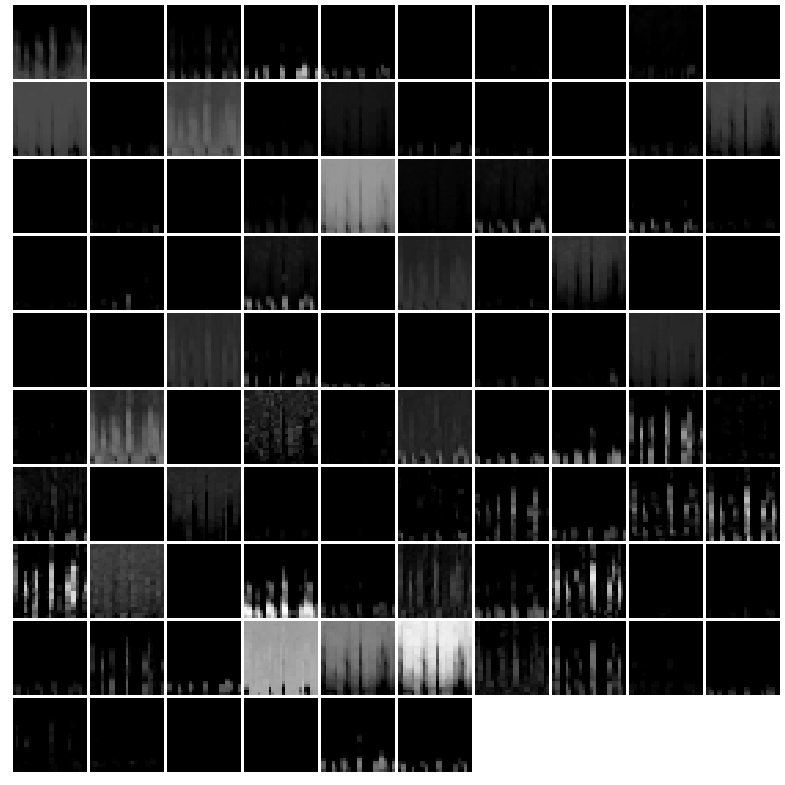}\\
\rotatebox[origin=c]{90}{Pool2} &\includegraphics[width=\linewidth]{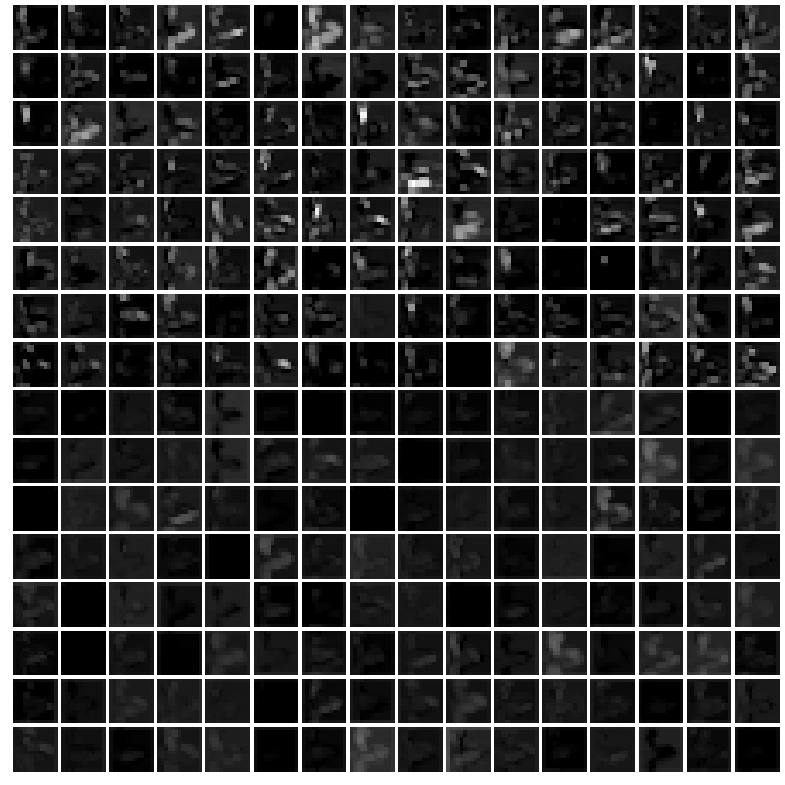} &
\includegraphics[width=\linewidth]{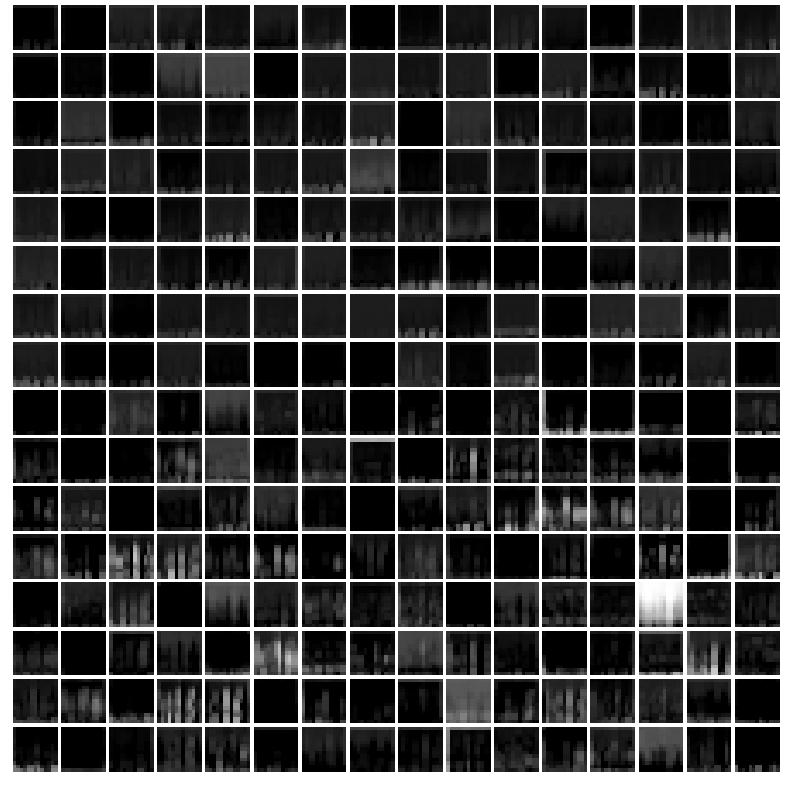} \\
\rotatebox[origin=c]{90}{Pool5} &\includegraphics[width=\linewidth]{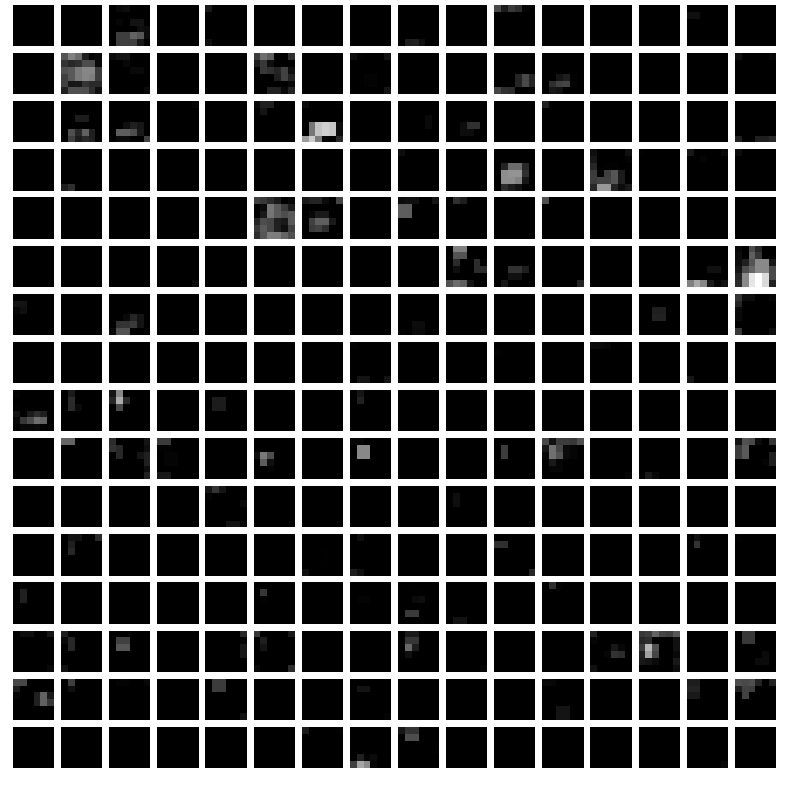} &
\includegraphics[width=\linewidth]{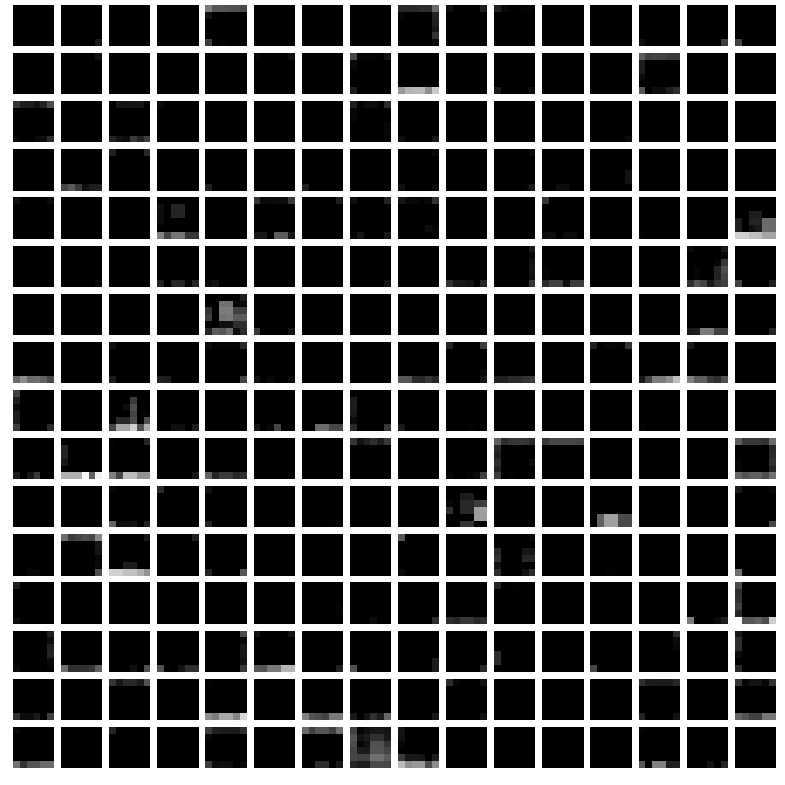} \\
\multicolumn{3}{c}{(a) Pooling layer}  \\
 & \multicolumn{2}{c}{\includegraphics[width=0.7\linewidth]{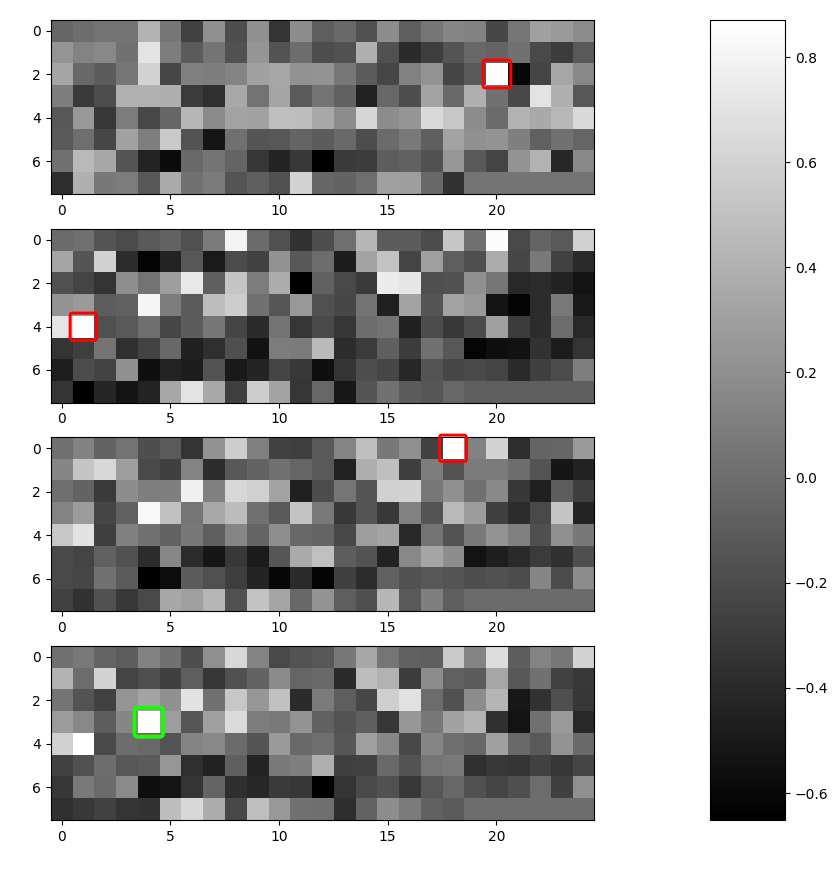}} \\
\multicolumn{3}{c}{(b) Fully connected layer} \\
\end{tabular}
\caption{Feature visualization of network layer. These are examples of features at different levels in Net3, where (a) shows the features of different pooling layer of image (left) and audio (right) network, and (b) top to bottom shows the features of last fully connected layer of image only, spectrogram only, Fused layer using FC7 concat, and fused layer using summation. Here, the red rectangle shows incorrect answer, the green rectangle shows correct answer of the classification.}
\label{activation_visualization}
\end{figure}
Another method of fine-tuning the model is to train Net3 in two stages. First, training the two stream individually followed by a joint fine-tuning. We train the image and audio CNNs separately, adapting the weights of pretrained model and learn the weights of the new 194-class output layer. After this training, the networks can be used to perform separate classification with respect to each modality. After then, we train an entire model by setting their learning rate to zero and only training the fusion part of the network to freeze the individual stream networks. As shown in Table \ref{table2}, two stage training is resulted best performance, and it proves that additional modality can improve the performance. We found that the fine-tuning pretrained model to two modalities and training them simultaneously, is significant worse than two stage training. We think the problem relates to the difference between the size of two datasets and batch size of single-modality and multi-modalities. In this experiment, we used net surgery\footnote{\url{https://github.com/BVLC/caffe/blob/master/examples/net_surgery.ipynb}} to fine-tune two different CNN using one pretrained model and to fine-tune two pretrained model into our model. 

To confirm that the features extracted from sample image and spectrogram during the fine-tuning were meaningful, we visualized the activations of different layers, especially the fused layers in Net\ref{net3}. The results shown in Fig.\ref{activation_visualization}, that allowed us to confirm the learned features were meaningful and qualitatively resembled the sample image and spectrogram. Moreover, the activations of our fused layer takes advantage by incorporating the features from each streams, when each network failed to produce write answer.  
\begin{table}[t]
	\caption{Classification performance of fine-tuned Net\ref{net3} model with different fusion and fine-tuning method.}
	\label{table2}
	\renewcommand{\arraystretch}{1.2}
	\centering
	\begin {tabular}{cl|c}
	\hline \hline%
	
	\multicolumn {2}{c}{Method} & \parbox{1.5cm}{\centering Accuracy (\%)}
	\\\hline
    \multicolumn {2}{c|}	{Fine-tuning weights of pretrained model (summation)}	& {65.0}	\\\hline
	\multirow{4}{*} {Two stage fine-tuning} 	& \textbf{Summation (ours, Net3)} 	& \textbf{78.9}		\\\cline {2-3}
	& Multiplication (ours, Net3) 	& 75.0		\\\cline {2 -3}
    & Simonyan et al. \cite{simonyan2014two}	& 70.0		\\\cline {2 -3}
	&  Eitel et al. \cite{eitel2015multimodal}	&  72.5	\\\hline
	\end {tabular}
\end{table}

\section{Conclusion}
\label{conclusion}
In this paper, we proposed three multimodal CNN architectures in different fusion strategies, which can process jointly the image and audio data for bird classification. Experimental results verified that the two-stream multimodal CNN in late fusion strategy outperforms the others. In addition, we proposed summed fusion method to combine multiple CNNs, which shows better performance comparing against several existing fusion methods. Moreover, with the help of two-stage fine-tuning, our method can be more effective. However, there still exist several drawbacks of our method: (1) Choosing the suitable duration based on the vocal features of birds to be recognized, is essential ingredients of improvement, is missed in our current work. (2) Our method is based on the raw image data, thus part detection and extracting features from pose-normalized regions may improve the classification performance.  
\par
As the future work, we aim to apply multiple kernel learning to combine multiple modalities, which is able to learn optimal composite kernel through combining basis kernels constructed from different features of modalities. We are also interested in designing new CNN architectures by increasing the number of modalities.


\bibliographystyle{ieicetr}
\bibliography{feature_combination}

%
%
%

\end{document}